\documentclass[11pt,a4paper]{article}
\usepackage{emnlp2022}
\usepackage{times}
\usepackage{latexsym}

\usepackage{times}
\usepackage{multicol}
\usepackage{latexsym}

\usepackage{enumitem,microtype}

\usepackage{amsmath}
\usepackage{amssymb}
\usepackage{graphicx}
\usepackage{caption, subcaption}
\usepackage{booktabs}
\usepackage{graphicx}
\usepackage{relsize}
\usepackage{latexsym}
\usepackage{mathabx}
\usepackage{xcolor}
\usepackage{xcolor,colortbl, makecell}
\usepackage{soul}

\usepackage{framed}
\usepackage{times}
\usepackage{url}
\usepackage{multirow}
\usepackage{mathtools}
\usepackage{enumitem}

\newcommand{\hlc}[2][yellow]{{%
    \colorlet{foo}{#1}%
    \sethlcolor{foo}\hl{#2}}%
}
\newcommand{\correctans}[1]{\hlc[green!30]{\textbf{#1}}}
\newcommand{\discourse}[1]{\hlc[cyan!30]{\textbf{#1}}}

\newcommand{\incans}[1]{\hlc[pink!80]{#1}}
\newcommand{\aquestion}[6]{\begin{tabular}{p{7.2cm}p{0.1cm}}
\multicolumn{2}{l}{\parbox{6.9cm}{\underline{#1} \\ #2 \\}} \\ 
 #3 &\\ 
 #4 &\\
 #5 &\\
 #6 &\\
\end{tabular}
}

\newlist{alphalist}{enumerate}{1}
\setlist[alphalist,1]{label=\textbf{\alph*.}}

\usepackage{microtype}
\newcommand{\comment}[1]{}

\title{\textsc{DiscoSense}: Commonsense Reasoning with Discourse Connectives}

\author{Prajjwal Bhargava \\
  University of Texas at Dallas \\
  \texttt{prajjwalin@protonmail.com} \\\And
  Vincent Ng \\
  University of Texas at Dallas  \\
  \texttt{vince@hlt.utdallas.edu} \\}

\date{}

\begin{document}
\maketitle

\begin{abstract}

We present \textsc{DiscoSense}, a benchmark for commonsense reasoning via understanding a wide variety of discourse connectives. We generate compelling distractors in \textsc{DiscoSense} using Conditional Adversarial Filtering, an extension of Adversarial Filtering that employs conditional generation. 
We show that state-of-the-art pre-trained language models struggle to perform well on \textsc{DiscoSense}, which makes this dataset ideal for evaluating next-generation commonsense reasoning systems.
\end{abstract}

\section{Introduction}
\label{sec:intro}

Much of the recent work in commonsense reasoning 
has focused on evaluating a pre-trained language model's (LM) ability to predict the most plausible ending/option given a context. Even after devising 
bias reduction techniques \cite{zellers-etal-2019-hellaswag, DBLP:conf/icml/BrasSBZPSC20} to mitigate the effects of annotation artifacts and make the task difficult, 
state-of-the-art LMs have managed to achieve or even surpass human performance on numerous commonsense downstream tasks \cite{zellers-etal-2019-hellaswag, Sakaguchi2020WINOGRANDEAA, bhagavatula2020abductive}.
Nevertheless, these LMs are still very far from being able to perform commonsense reasoning as well as humans. 
Hence, the fact that they have begun to ace existing benchmarks implies that time is ripe to design a new challenging benchmark that can reliably target their limitations.

Motivated by this observation, we present \textsc{DiscoSense}, a benchmark for performing commonsense reasoning through understanding a wide variety of discourse connectives. Figure~\ref{tab:teaser_examples} shows an example taken from \textsc{DiscoSense}. As can be seen, an example is composed of a {\em context} (e.g., ``Our waitress was very nice, but she kept on forgetting my stuff.") and a discourse connective (e.g., ``For example"), and the goal is to choose the most plausible ending out of four options.
If we {\em ignore} the discourse connective, then all four options may seem plausible because we do not know what the writer's intent is.
Once we consider both the context and the discourse connective, then it is clear that only option b) is plausible. The reason is that 
``For example" signals an {\sc Exemplification} relation between its arguments, and what follows the discourse connective is expected to be an example of the waitress keeping on forgetting the writer's stuff. 
Using commonsense knowledge, we know that (1) ``my beer and fries" is an example of ``my stuff", and (2) her taking forever to bring the writer stuff implies she kept on forgetting his/her stuff.


\begin{figure}[t]
\centering\small
    {\FrameSep1pt
    \begin{framed}\footnotesize \begin{tabular}{@{} p{7cm}} 
 Our waitress was very nice, but she kept on forgetting my stuff. \discourse{For example} \\
 \vspace{1mm}
a)  When I ordered the garlic shrimp, she remembered to\\ \hspace{3mm} add my requested garlic butter.\\
b)  {\bf She took forever to bring me my beer and fries.}\\
c)  When I told her I wanted to use the free breakfast that\\ \hspace{3mm} was available she was not pleased.\\
d)  For some customers, this is fine.\\
\end{tabular}
\end{framed} }

\caption{Example on commonsense reasoning with discourse connectives. The correct (i.e., most plausible) option is boldfaced.} 
\label{tab:teaser_examples}
\vspace{-3mm}
\end{figure}

What if we replace ``For example" with ``However" in the example? 
Since ``However" signals a {\sc Contrast} relation, 
options a) and d) both seem viable. Specifically, option a) describes a situation in which she did not forget the writer's stuff. While option d), unlike option a), does not describe any example that signals a contrast, one may infer a contrast between option d) and the context: being forgetful is fine for some customers. Nevertheless, option a) is arguably {\em more plausible} than option d) and should be chosen.
The reason is that for d) to be sensible, one needs to assume that her forgetting {\em the writer}'s stuff implies that she is in general forgetful. Without this assumption, it may be strange for other customers to have an opinion on her forgetting the writer's stuff. In general, the most plausible option is the option that 
makes the smallest number of assumptions, and/or is the most coherent given the context and the discourse connective.
Considering the commonsense knowledge {\em and} the reasoning involved, it should not be difficult to see that this task is challenging.

Our contributions are four-fold. First, we create \textsc{DiscoSense}, a new dataset aimed at testing 
LMs' commonsense reasoning capabilities through discourse connectives. Second, we employ a controlled text generation based adversarial filtering approach to generate compelling negatives. Third, we establish baseline results on \textsc{DiscoSense} with numerous state-of-the-art discriminator models and show that they struggle to perform well on \textsc{DiscoSense}, which makes our dataset an ideal benchmark for next-generation commonsense reasoning systems.
Finally, we show the efficacy of using \textsc{DiscoSense} as a transfer learning resource through  sequential fine-tuning of LMs on \textsc{DiscoSense} followed by \textsc{HellaSwag} and achieve near state-of-the-art results on the \textsc{HellaSwag} test set. 
To stimulate work on this task, we make our code and data publicly available.%
\footnote{For our code and data, see \url{https://github.com/prajjwal1/discosense/}.}

\comment{
Unlike these benchmarks, \textsc{DiscoSense} 
requires 
reasoning over discourse relations. 
In addition, unlike \textsc{Swag} and \textsc{HellaSwag} where examples come primarily from ActivityNet (a benchmark focused primarily on dense captioning of temporal events), \textsc{DiscoSense} features a  more diverse set of examples coming from varied domains that may only be solved with rich background world knowledge.
}

\section{Related Work}
\label{sec:related}

In this section, we discuss related work, focusing our discussion on the differences between {\sc DiscoSense} and existing commonsense reasoning benchmarks. In addition, we present an overview of Adversarial Filtering, which will facilitate the introduction of the Conditional Adversarial Filtering mechanism we propose in Section~\ref{sec:DiscoSense}.

\paragraph{Commonsense reasoning benchmarks.} \textsc{Swag} \cite{zellers-etal-2018-swag} and \textsc{HellaSwag} \cite{zellers-etal-2019-hellaswag} are arguably the most prominent commonsense reasoning benchmarks. In \textsc{Swag}, given a partial description along with four candidate endings, the task is to predict the most plausible ending. The synthetic options (a.k.a.\ distractors) are generated through a process called Adversarial Filtering (AF) (see below). 
\textsc{HellaSwag} is an extension of \textsc{Swag} that seeks to eliminate artifacts in the generated endings.
Unlike \textsc{Swag} and \textsc{HellaSwag}, \textsc{DiscoSense} requires that the discourse connective be taken into account in the reasoning process, thus increasing the number of inference steps and potentially the task complexity. In addition,
while the examples in \textsc{Swag} and \textsc{HellaSwag} come primarily from ActivityNet (a benchmark focused on dense captioning of temporal events), \textsc{DiscoSense} features a  more diverse set of examples coming from varied domains that may only be solved with rich background knowledge.

There are 
benchmarks that aim to test different kinds of commonsense reasoning abilities, although none of them focuses on reasoning over discourse connectives. 
SocialIQA \cite{sap-etal-2019-social}, for instance, 
focuses on social and emotional commonsense reasoning. \textsc{Abductive NLI} \cite{bhagavatula2020abductive} focuses on abductive reasoning.
\textsc{Winogrande} \cite{Sakaguchi2020WINOGRANDEAA} contains Winograd schema-inspired problems, which are essentially hard pronoun resolution problems requiring world knowledge. 
PIQA \cite{Bisk2020} examines physical commonsense reasoning. 
\textsc{Mc-Taco} \cite{zhou-etal-2019-going} and \textsc{TimeDial} \cite{qin-etal-2021-timedial} 
focus on temporal reasoning in comprehension and dialogue formats. 

More closely related to {\sc DiscoSense} are commonsense reasoning benchmarks that involve reasoning with 
a particular kind of relations. 
{\sc COPA} (Choice of Plausible Alternatives) \cite{copa} focuses exclusively on reasoning with {\sc Causal} relations and involves choosing the more plausible ending out of two (rather than four) options.
{\sc P-MCQA} \cite{DBLP:journals/corr/abs-2104-08712} focuses exclusively on reasoning with {\sc Precondition} relations: given a commonsense fact, select the precondition that make the fact possible (enabling) or impossible (disabling)
out of four options.
{\sc $\delta$-NLI} \cite{rudinger-etal-2020-thinking}, which aims to evaluate {\em defensible inference}, focuses exclusively on reasoning with the {\sc Strengthen}/{\sc Weaken} relations: given a premise-claim pair where the premise supports the claim, generate a sentence that either strengthens or weakens the support. 
{\sc WinoVenti} \cite{do-pavlick-2021-rotten}, which is composed of Winograd-style schemas, focuses exclusively on reasoning with {\sc Entailment} relations: given two sentences with an entailment relation, such as "Pete says the pear is delicious. The pear is \_\_\_\_", the goal is to fill in the blank with one of two choices (e.g., "edible", "inedible").
There are two key differences between these datasets and {\sc DiscoSense}.
First, rather than focusing on a particular type of relation, {\sc DiscoSense} encompasses 37 discourse connectives signaling different discourse relation types. Second, {\sc DiscoSense} involves reasoning with discourse {\em connectives}, which is more complicated than reasoning with discourse relations. Specifically, as some 
connectives are sense-ambiguous (e.g., the connective "since" may serve as a temporal or causal connective \cite{pitler:acl09s}), a LM will likely need to (implicitly) perform sense disambiguation in order to perform well on {\sc DiscoSense}.

\begin{table}[t!]
\begin{small}
\centering
\setlength{\tabcolsep}{2.5pt}
\begin{tabular}{lcccc}
  \toprule
  Dataset & Model & Human \\
  \midrule
  SWAG \cite{zellers-etal-2018-swag} & 91.71 & 88\\
  $\alpha$NLI \cite{bhagavatula2020abductive} & 91.18 & 92.9\\
  Hellaswag \cite{zellers-etal-2019-hellaswag} & 93.85 & 95.6 \\
  CosmosQA \cite{huang-etal-2019-cosmos} & 91.79 & 94\\
  PIQA \cite{Bisk2020} & 90.13 & 94.9\\
  SocialIQa \cite{sap-etal-2019-social} & 83.15 & 88.1\\
   MC-TACO  \cite{zhou-etal-2019-going} & 80.87 & 75.8\\
  WinoGrande \cite{Sakaguchi2020WINOGRANDEAA} & 86.64 & 94 \\
  ProtoQA \cite{boratko-etal-2020-protoqa} & 54.15 & 74.03\\
  VCR \cite{zellers2019vcr} & 63.15 & 85\\
  \bottomrule
\end{tabular}
  \caption{
    Status of how competitive current commonsense reasoning benchmarks are for state-of-the-art pre-trained language models. 
  }
\label{tab:status_difficulty_benchmark}
\end{small}
\vspace{-2mm}
\end{table}

There are datasets and knowledge bases where the semantic/discourse/commonsense relations are explicitly annotated and which can provide data sources from which commonsense reasoning benchmarks can be derived.
Examples include (1) the Penn Discourse TreeBank \cite{prasad:lrec08}, where two sentences or text segments are annotated with their discourse relation type, if any; (2) {\sc CoreQuisite} \cite{DBLP:journals/corr/abs-2104-08712}, which is used to provide the commonsense facts and the human-generated preconditions in the {\sc P-MCQA} dataset mentioned above;
(3) {\sc SNLI} \cite{bowman:emnlp15}, where each premise-hypothesis pair is annotated as {\sc Entailment}, {\sc Contradiction}, or {\sc Neutral}; (4) {\sc ATOMIC}$^{20}_{20}$ \cite{Hwang2021COMETATOMIC2O}, which is a commonsense knowledge graph where the nodes correspond to propositions and the edges correspond to social/physical commonsense relations; and (5) {\sc SOCIAL-CHEM-101} \cite{forbes:emnlp20}, which is a collection of statements about commonsense social judgments made given everyday situations.

One of the motivations behind the creation of \textsc{DiscoSense} is that state-of-the-art LMs have managed to achieve or even surpass human performance on various commonsense reasoning benchmarks. 
Table~\ref{tab:status_difficulty_benchmark} shows the best accuracies achieved by existing LMs on 10 widely used commonsense reasoning benchmarks and the corresponding human performance levels. As can be seen, existing LMs have managed to achieve an accuracy of more than 80\% on eight of these benchmarks. 

\begin{figure}[!t]
\centering
    \includegraphics[height=0.43\linewidth, scale=0.2]{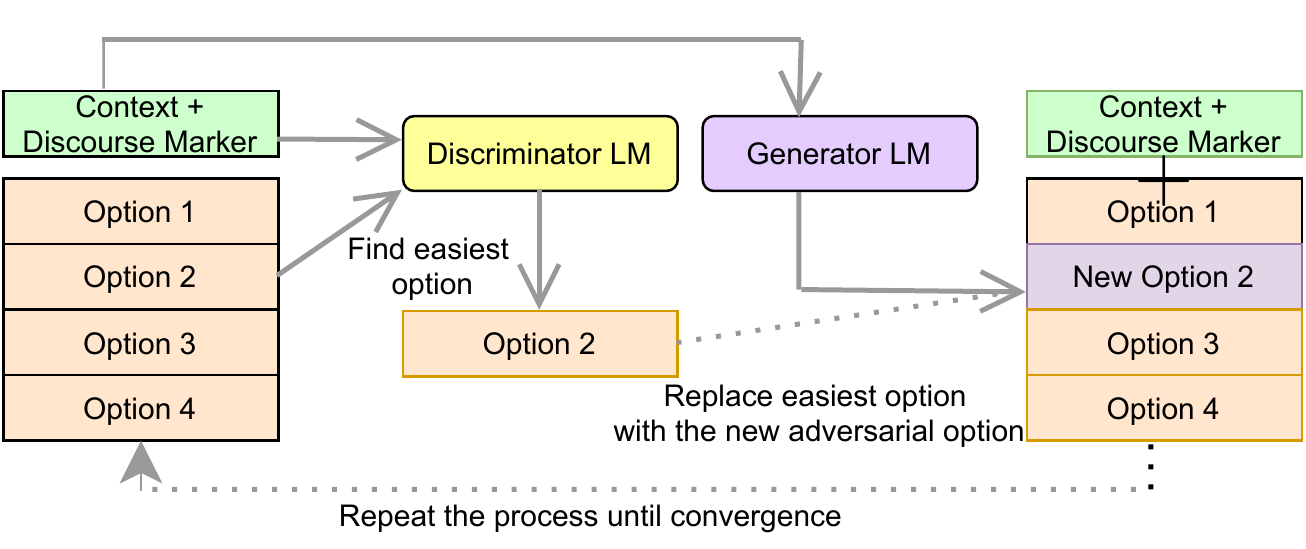}
  \caption{Components of Adversarial Filtering.}
\label{fig:af}
\vspace{-2mm}
\end{figure}

\paragraph{Adversarial filtering (AF).} 
Originally proposed by \newcite{zellers-etal-2018-swag}, AF aims to create examples that would be difficult for models to solve, specifically by replacing the easy options in correctly-solved examples with difficult ones. 
As shown in ~\autoref{fig:af}, AF has three components: data (i.e., examples with multiple options, one of which is correct), a discriminator LM (a classifier that is used to solve each example) and a generator LM (a model that generates new options for an example). 
In each AF iteration, the discriminator LM is trained on the training set and used to solve each example in the test set. 
If a test example is incorrectly solved (i.e., the discriminator LM chooses the wrong option), the example is deemed sufficiently difficult and no change is made to it. On the other hand, if a test example is correctly solved, then AF seeks to increase its difficulty by replacing the {\em easiest} option (i.e., the generated option that the discriminator LM classifies with the highest confidence)
with a new option generated by the generator LM.
Training a new discriminator LM in each AF iteration 
ensures that the dataset is not just adversarial for one LM but a class of LMs, as
training different instances of the same type of LMs results in models that have differently learned linguistic representations. 
This process is repeated on all correctly classified examples in the test set until the performance on the test set converges. 


\section{\textsc{DiscoSense}}
\label{sec:DiscoSense}
\subsection{Task Description}
\textsc{DiscoSense} aims to measure the commonsense inference abilities of computational models through the use of discourse connectives. The correct endings can be obtained after understanding the purpose of the given discourse connectives. Given a context $c = (s, d)$, which is composed of a contextual sentence $s$ and a discourse connective $d$ as well as a set of four options $O=\{o_{1}, o_{2}, o_{3}, o_{4}\}$, the task is to predict the most plausible ending $o_{i} \in O$.
\subsection{Dataset Creation}
\label{sec:dataset_creation}

To assemble \textsc{DiscoSense},
we focus on source datasets that contain two sentences connected through a discourse connective. 
Specifically,
we use two peer reviewed academic datasets, \textsc{Discovery}~\cite{sileo-etal-2019-mining} and \textsc{Discofuse} ~\cite{geva-etal-2019-discofuse}.
In \textsc{Discovery}, each sentence is composed of two sentences connected via a discourse connective for the purpose of learning joint sentence representations with discourse connectives. 
\textsc{Discofuse}, on the other hand, is assembled for the task of sentence fusion (i.e., joining several independent sentences into a single coherent sentence). We only consider those examples 
where a discourse connective is needed for sentence fusion, and include in \textsc{DiscoSense} the fused sentences in the Wikipedia%
\footnote{\url{https://en.wikipedia.org/}} split of \textsc{Discofuse}.   
Since these datasets contain sentences from Common Crawl%
  \footnote{\url{https://commoncrawl.org/}} and Wikipedia
   articles, \textsc{DiscoSense} is diverse in the topics it covers. Importantly, since by construction the discourse connective is crucial in solving the underlying tasks (i.e., sentence representation learning and sentence fusion), the crucial role played by the discourse connectives in these sentences makes them suitable for our use case.
Details of how the {\sc Discovery} and {\sc Discofuse} sentences are used to create {\sc DiscoSense} are shown in Tables~\ref{tab:DiscoSense_data_split} and~\ref{tab:generator_data}.

\comment{
    \textbf{\textsc{Discovery}} 
    We used this dataset in its entirety. It has been collected from Depcc ~\cite{panchenko-etal-2018-building} corpus, which consists of English text harvested from commoncrawl web data.
    \textbf{\textsc{DiscoFuse}} primarily targets the task of sentence fusion defined as task of joining several independent sentences into a single coherent text. During data curation process of \textsc{Discofuse}, texts were selected on the basis of detected phenomenon occuring within them such as discourse/forward/inner connective, anaphora, cataphora, clause and coordination. We only select those examples that are concerned with discourse connective phenomena since it is required that sentences have to be fused necessarily with a discourse connective to form a coherent fused sentence.  In order to increase concept coverage and diversity, we use the Wikipedia split amongst Wikipedia and Sports splits. 
 }


\begin{table} [t]
\begin{center}
\footnotesize{
\begin{tabular}{lrr}
\toprule
\multicolumn{1}{c}{Data} & \multicolumn{1}{c}{\textsc{DiscoSense}}  & \multicolumn{1}{c}{\textsc{DiscoSense} } \\
\multicolumn{1}{c}{Source} & \multicolumn{1}{c}{Train}  & \multicolumn{1}{c}{Test} \\
\midrule
\textsc{Discovery} Train & Bottom 7\% & - \\
\midrule
\textsc{Discovery} Validation & - & 100\% \\
\bottomrule
\textsc{Discofuse} train & Top $\sim$54k  & - \\
                         & w/ DC & \\
\hline
\end{tabular}}
\end{center}
\caption{
Data sources for 
\textsc{DiscoSense} and its composition before human verification. DC refers to those samples in \textsc{Discofuse} that are concerned with the discourse connective phenomenon.}
\label{tab:DiscoSense_data_split}
\vspace{-2mm}
\end{table}

\comment{
\subsubsection{Background} 
Typically when LMs are asked to generate samples from LM iteratively by sampling next token, we do not have control over attributes of generated text such as style, sentiment etc. However there are many use cases where we require good control over output text. For example, if we are generating content for public, we want it to be safe and informative for people to read. We now look at approaches who have addressed this requirement of controlling attributes of generated text.

Controllable Text Generation aims to provide a more granular control over how generation happens to match a particular attribute. In the context of transformer based PLMs, previous works have looked at how fine-tuning extra set of parameters while keeping the base model (unconditionally trained) fixed to steer generation \cite{Dathathri2020Plug, qin-etal-2020-back, 10.1007/978-3-030-58580-8_41, KrauseGeDi2020}
and conditionally training a generative model on a control variable to generate text with respect to a prompt prefix \cite{keskar2019ctrl}. We extend the later approach of \textsc{Ctrl} \cite{keskar2019ctrl} to explicitly steer generation with respect to discourse relations. Using discourse connective as control codes helps generate sentences that predominantly comply with the purpose of the connective. Our quantitative and qualitative results show that this approach provides improvements over traditional conditional generation mechanism and makes the generation relatively more human-like.
}

\subsection{Generating Options}

Next, we describe how we generate challenging options for \textsc{DiscoSense} using an improved version of AF that we call Conditional Adversarial Filtering (CAF). CAF follows the AF procedure in Figure~\ref{fig:af}, only differing from AF in terms of (1) the generator LM (Section~3.3.1), (2) the discriminator LM (Section~3.3.2), and (3) how the generator LMs are used to generate options (Section~3.3.3). 

\subsubsection{Conditional Generator LM}
Pre-training does not explicitly teach how important a particular token or text span is in contributing to the semantics of a sentence. Hence, to be able to generate sentences that are coherent with not only the context but also the discourse connective, 
we propose to use Controllable Text Generation, which aims to provide a more granular control over how generation happens to match a particular attribute. In the context of Transformer-based LMs, there are two lines of research on controllable text generation. One examines how to steer generation by fine-tuning an extra set of parameters while keeping the base (unconditionally trained) model fixed \cite{Dathathri2020Plug, qin-etal-2020-back, 10.1007/978-3-030-58580-8_41, KrauseGeDi2020}, while the other involves conditionally training a generative model on a control variable to generate text w.r.t.\ a prompt prefix. 
We adopt the latter approach, extending \textsc{Ctrl} \cite{keskar2019ctrl} to explicitly steer generation w.r.t.\ discourse relations by using discourse connectives as control codes, as described below.



\vspace{1mm}
\textbf{Training.}
The input to CTRL is as follows:

\vspace{-6mm}
\begin{align*}
\begin{split}
     \text{\underline{input}: } \; [d]+[\text{context}] - \text{\underline{label}: } \; [\text{ending}] \\
\end{split}
\end{align*}
\vspace{-6mm}

{\noindent where $d$ is a discourse connective. Specifically, each
input context for \textsc{Ctrl} is prepended with a 
connective, and the training task for \textsc{Ctrl} is to learn the conditional distribution $p(e|d,{\rm context})$ over possible endings $e$. The predicted ending is then compared with the human generated ending to compute loss. Since the original \textsc{Ctrl} model is pre-trained with control codes suitable for open-ended text generation, we fine-tune \textsc{Ctrl} on the portion of \textsc{Discovery} shown in \autoref{tab:generator_data} using all the 174 connectives present in the selected splits.
Comparing Tables~\ref{tab:DiscoSense_data_split} and~\ref{tab:generator_data}, we can see that the data the generator LM is fine-tuned on is not part of \textsc{DiscoSense}. Doing so ensures that the endings generated by the generator LM are different from the ground truth (i.e., the human written endings).}


\comment{
\begin{table}[t]
\footnotesize
\centering
\begin{tabular}{lccc}
\toprule
Generator & Perplexity \\ 
\midrule
\textsc{GPT2-XL} & 2.53 \\
\textsc{CTRL} & 2.39 \\
\bottomrule
\end{tabular}
\caption{Perplexity scores obtained with generator LMs on the Discovery validation set. Lower is better.}
\vspace{-2mm}
\label{tab:perplexity}
\end{table}
}

\begin{table} [t]
\begin{center}
\footnotesize{
\begin{tabular}{lr}
\toprule
\multicolumn{1}{c}{Data} & \multicolumn{1}{c}{Generator LM}  \\
\midrule
{\sc Discovery} Train & last 93\% \\
{\sc Discovery} Test & 100\%  \\
\bottomrule
\end{tabular}}
\end{center}
\caption{Data used to train the generator LMs in Conditional Adversarial Filtering.}
\label{tab:generator_data}
\vspace{-2mm}
\end{table}

\textbf{Decoding.} We use Nucleus sampling \cite{Holtzman2020The} for generating options for the training set with the value of $p$ set to $0.7$, which means the weights of the tail of the probability distribution are ignored (i.e., tokens with a cumulative probability mass of less than $0.3$ are left out). 
Additionally, 
we use a length penalty of $0.8$ to restrict the length of the generations to match the average length of the ground truth to avoid the induction of length bias.

\comment{
\textbf{Decoding:} We use Nucleus sampling \cite{Holtzman2020The} for generating options for the training set with the value of $p$ set to $0.7$, which means probability weights of the tail are ignored (tokens with commulative probability mass $<$ $0.3$). To generate the test set, we change $p$ to $0.98$ in order to increase diversity (i.e., the distribution gap between training and test). 
Additionally, we use a length penalty of $0.8$ to restrict the length of generations to match the average lengths of ground truth to avoid induction of length bias.
}

\textbf{Efficacy of conditional generation.} Recall that we propose the use of conditional generation, specifically the use of discourse connectives as control codes, in our generator LM because of our hypothesis that the resulting LM would generate options that are more compliant with the purpose of the discourse connective. To test this hypothesis, we compare the {\em text generation} capability of \textsc{Ctrl} with that of \textsc{Gpt2-xl}, a model that is trained unconditionally and has nearly the same number of parameters (1.6B) as \textsc{Ctrl}, under the {\em same} evaluation setting. Specifically, both LMs are fine-tuned on the same data (see Table~\ref{tab:generator_data}) using the same machine (a 2x Quadro RTX 8000 with a batch size of 24). The only difference between them lies in the format of the training examples: in \textsc{Ctrl} the discourse connective is used as the control code and therefore precedes the context, whereas in \textsc{Gpt2-xl}, the discourse connective follows the context. 

The two LMs are then independently applied to generate exactly one option for each example in the \textsc{Discovery} validation set. 
\textsc{Ctrl} achieves a much lower perplexity than \textsc{Gpt2-xl} (2.39 vs.\ 2.53), which suggests
that conditional training improves the quality of the generated sentences.

\comment{
In order to validate if conditional generation mechanism is beneficial for our task, we compare performance of \textsc{CTRL} with \textsc{GPT2-XL} that is trained unconditionally and nearly has the same parameter count (1.6B). We fine-tuned this pre-trained variant of \textsc{GPT} on the same task as that of \textsc{CTRL} i.e given a context followed by a discourse connective, it is evaluated on its ability predict a single ending. Training is performed on the same exact data and experimental setting. Note that we are interested in examining these models for their text generation capability and not how they would perform when coupled in CAF. From ~\autoref{tab:perplexity}, we show that there is a significant performance difference in text generation between our model and GPT2-XL, validating that conditional training does provide text generation improvements over the standard way.
}

\subsubsection{Discriminator LM}
We use \textsc{Roberta-large} \cite{liu2019roberta} as the 
discriminator LM, which 
takes the context, the discourse connective, and the four endings as input and predicts the most plausible ending. This LM is trained on the randomly shuffled training split of \textsc{DiscoSense} 
and applied to the \textsc{DiscoSense} test set 
to get the confidence scores associated with its predictions.


\subsubsection{Generating Options}

Next, we describe how we generate options for the examples in 
\textsc{DiscoSense}. Recall that each example contains one of 174 discourse connectives. Rather than generating options for examples that contain any of these 174 connectives, we select 37 discourse connectives and generate options only for examples that contain one of them. The connectives that are discarded are primarily those that impose few constraints on the endings to be generated given the context according to preliminary experiments. For instance, the connective ``and" is discarded because numerous endings are equally plausible. Similarly for connectives that signal a temporal relation (e.g., ``before", ``after"): they also tend to allow numerous equally plausible endings, as can be seen in examples such as ``John went to eat lunch after [ending]". The 37 connectives that we end up choosing are shown in Table~\ref{tab:markers_list}. These connectives are less likely to yield options that look equally plausible to human annotators 
and which are indicative of different kinds of discourse relations, such as {\sc Exemplification} (e.g., ``for instance"), {\sc Concession} (e.g., ``although"), {\sc Comparison} (e.g., ``in contrast"), and {\sc Causal} (e.g., ``as a result"). 94k examples in  \textsc{DiscoSense} 
contain one of the 37 connectives. 

\begin{table}[t!]
\begin{center}
\begin{small}
\begin{tabular}{lll}
\toprule
although & in other words & particularly\\
as a result &  in particular & rather\\
by contrast & in short & similarly\\
because of this & in sum & specifically\\
because of that & interestingly & subsequently\\
but & instead & thereafter\\
consequently & likewise & thereby\\
conversely & nevertheless & therefore\\
for example & nonetheless & though\\
for instance & on the contrary & thus\\
hence & on the other hand & yet\\
however & otherwise &\\
in contrast & overall & \\
\bottomrule
\end{tabular}
\end{small}
\end{center}
\caption{Discourse connectives present in \textsc{DiscoSense}.}
\label{tab:markers_list}
\vspace{-2mm}
\end{table}

To generate the options for these 94k sentences,
we begin by training 20 generator LMs
on a randomly shuffled order of the generators' training data (see Table~\ref{tab:generator_data}) and then inserting them into a circular queue. Although the underlying data is the same, random shuffling ensures that the learned representations of these 20 models are different. 
Since each example needs to have 3 synthetic options, we use the first 3 generator LMs from the circular queue to generate the initial options for each example. After that, we begin CAF. In each CAF iteration, we (1) train the discriminator LM (see Section~3.3.2) on the \textsc{DiscoSense} training set for $4$ epochs and use it to filter out the options deemed as easiest by the discriminator LM; and (2) use the next generator LM in the circular queue to generate the options for the examples whose easiest option is removed by the discriminator LM. In other words, a different discriminator LM is used in each CAF iteration, and a generator LM in the circular queue is used once every 20 CAF iterations.
CAF is run separately for the \textsc{DiscoSense} training and test sets.
After running CAF for approximately 150 iterations, the average accuracy of a discriminator LM decreased from 86--90$\%$ to 34$\%$ on the \textsc{DiscoSense} test set.

\begin{table}[t]
   \footnotesize
    \centering
    \begin{tabular}{@{}llr@{}}
    \toprule
\multicolumn{3}{@{}c@{}}{DiscoSense} \\ \midrule
    \multirow{4}{2.3cm}{\# Context Answer tuples} & train &  9299 \\ 
    & test & 3757 \\
    & \textbf{total} & \textbf{13056} \\\bottomrule
    \toprule
    \multicolumn{3}{@{}c@{}@{}}{Statistics}{Train / Test} \\ \midrule
    \multirow{5}{0.1cm}{Average \\\# tokens} & context & 22.08 / 22.51 \\
        & answers (all) & 18.62 / 18.92 \\
        & answers (correct) & 16.94 / 18.18 \\
        & answers (incorrect) &  18.51 / 18.5 \\ \midrule
    \multirow{4}{0.1cm}{Unique \\\# tokens} & context & 32577 / 16858\\
        & answers (all) &  43992 / 27406\\
        & answers (correct) & 26836 / 15078 \\
        & answers (incorrect) & 41158 / 25900 \\ \midrule
    \bottomrule
    \end{tabular}
    \caption{Data statistics for \textsc{DiscoSense}.}
    \label{tab:data-stats}
    \vspace{-2mm}
\end{table}

\subsubsection{Other Implementation Details}

 For the models we use in CAF, we obtain the pre-trained weights and the 
 implementations from Hugging Face Transformers ~\cite{DBLP:journals/corr/abs-1910-03771}. These models are trained using the AdamW optimizer~\cite{loshchilov2018decoupled} with a learning rate of $2e^{-5}$. 
 The training of each generator LM is performed on a $2$x Quadro RTX 8000 with a batch size of 24 and typically lasts for 3 days. The training 
 of a discriminator LM is performed on a RTX $3090$ with a batch size of 16 and typically 
 lasts for 5--6 hours.

\subsection{Human Verification}

Next, we perform human verification of the examples for which we have generated options. The verification proceeds in two steps.
In Step~1, we ask three human verifiers to independently identify the correct option for each example, removing an example if at least one person fails to identify the correct option. We repeat this process until the number of examples that survive this verification reaches 13,056.%
\footnote{This is the maximum number we can handle given our budgetary constraints.}
In Step~2, we ask three human verifiers not involved in Step~1 to independently identify the correct option for each of the 13,056 examples verified in Step~1. We compute for each verifier the accuracy of choosing the correct option and use the average accuracy as the human performance on {\sc DiscoSense}.
Appendix~A contains the details on how the human verifiers are recruited and the annotation instructions we present to them.

\subsection{Dataset Statistics}

Statistics on \textsc{DiscoSense} are shown in Table~\ref{tab:data-stats}, in which we 
report the average number of tokens in (1) the context, (2) the ground truth and (3) the generated endings. The number of unique tokens provides a rough characterization of the richness of the vocabulary. 
In addition, we report the distribution of the examples over the discourse connectives 
in \textsc{DiscoSense} in Figure~\ref{fig:training_connective_stats}.

\comment{
\begin{figure}[!t]
\centering
    \includegraphics[height=1.5\linewidth, scale=0.18]{disco_sense/figs/connective_stats.pdf}
  \caption{Distribution of discourse connective across \textsc{DiscoSense}.}
\label{fig:training_connective_stats}
\end{figure}
}

\begin{figure}[!t]
\centering
    \includegraphics[height=0.8\linewidth, scale=0.14]{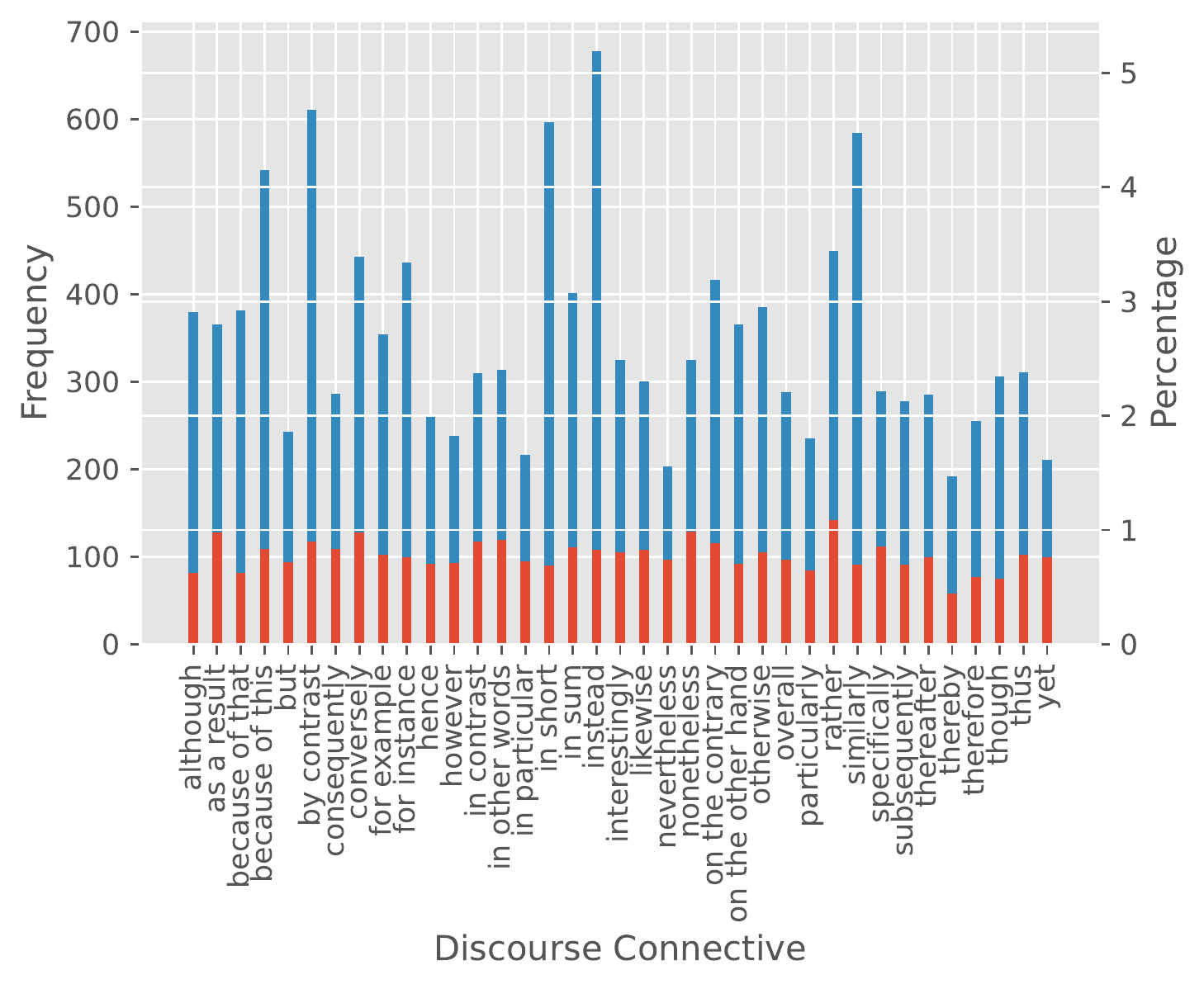}
  \caption{Distribution of examples over discourse connectives in \textsc{DiscoSense}.}
\label{fig:training_connective_stats}
\vspace{-2mm}
\end{figure}

\section{Evaluation}
\label{sec:experiments}


\subsection{Baseline Systems}

Our baselines are composed of prominent LMs with
different kinds of Transformer architectures.
First, we consider models that are pre-trained in a \textsc{Bert}-like fashion and share architectural similarities, including the base and large variants of \textsc{Bert} \cite{devlin-etal-2019-bert} and \textsc{Roberta} \cite{liu2019roberta}, as well as \textsc{Albert-xxlarge-V2} \cite{Lan2020ALBERT:}. As an extension, we select \textsc{Longformer Base}, which is pre-trained in the same manner as \textsc{Roberta} but has a sparse attention matrix.%
\footnote{Some endings are longer than the others. The use of \textsc{Longformer} allows us to see whether a sparse attention matrix can better exploit the length of an ending than other models.}
From the autoregressive/decoder based networks, we experiment with \textsc{XLNet Large} \cite{NEURIPS2019_dc6a7e65}, which maximizes the learning of bidirectional contexts and \textsc{Gpt2-XL}. For models trained with a different pre-training objective, we experiment with \textsc{Electra-Large} \cite{Clark2020ELECTRA:} and \textsc{Funnel-Transformer-XL} \cite{DBLP:conf/nips/DaiLY020}, the latter of which is pre-trained in a similar manner as \textsc{Electra-Large}.

We obtain the implementations of these LMs from Hugging Face Transformers. We fine-tune them on the \textsc{DiscoSense} training set using a 4-way cross-entropy loss in the same way as the discriminator LMs in CAF are trained (see Section 3.3.4) 
and evaluate them on the test set.

\comment{
The implementations of these PLMs are obtained
from Huggingface Transformers \cite{DBLP:journals/corr/abs-1910-03771}. 
All 
PLMs are fine-tuned on the \textsc{DiscoSense} training set using a 4-way cross-entropy loss in the same way as the discriminator LMs in CAF were trained (see Section 3.3.4 for details), and are subsequently evaluated on the test set. 
}

\begin{table}[t]
\footnotesize
\begin{tabular}{lccc}
\toprule
\bf Model & \bf Accuracy / std \\ 
\midrule
Random Guess & 25.0\\
\textsc{Bert-Base} (110M) & 32.86 / 0.45\\
\textsc{Bert-Large} (336M) & 34.25 / 1.04\\
\textsc{Roberta-Base} (125M) & 34.11 / 0.45\\
\textsc{Roberta-Large} (355M) &  34 / 0.2\\
\textsc{Albert-xxlarge-V2} (223M) & 50.91 / 1.44 \\
\textsc{Longformer Base} (435M) & 35.29 / 0.77\\
\textsc{XLNet Large} (340M) & 36.71 / 0.77 \\
\textsc{Funnel-Transformer-XL} (468M) & 35.22 / 1.94 \\
\textsc{Electra-Large} & 65.87 / 2.26\\
Human Performance & 95.40 / 0.20\\
\bottomrule
\end{tabular}
\caption{Accuracies (best results obtained among 8 epochs when averaged over 5 runs with random seeds) of the LMs on the \textsc{DiscoSense} test set.}
\label{tab:quantitative_main_results}
\vspace{-2mm}
\end{table}

\subsection{Results and Discussion}
Results on the test set, which are expressed in terms of accuracy, are shown in \autoref{tab:quantitative_main_results}. A few points deserve mention.

First, all baselines perform better than random guess (row~1). 
This implies that while CAF is used 
to remove easy options, there may still be artifacts in the data that could be exploited by the LMs.

Second, models sharing a similar pre-training objective as that of \textsc{Bert}, such as \textsc{Roberta} and \textsc{Longformer}, are among the worst baselines.
A similar trend is observed with \textsc{XLNet}. Although \textsc{Albert} has the Masked Token Prediction task in its pre-training objective, its architectural differences (i.e., larger hidden states and parameter sharing) and its Sentence Order Prediction objective seem to help it learn inter-sentence coherency properties better than its \textsc{Bert} counterparts.

Third, pre-training appears to play a predominant role in our task. 
While the \textsc{Bert} family of models are trained with the masked-LM objective, 
the pre-training objective of \textsc{Electra} (the best baseline) is designed to determine if a token in a human-written sentence has been replaced by a generator. 
We speculate that \textsc{Electra}'s superior performance can be attributed to the fact that its pre-trained knowledge of discriminating between synthetic and human generated tokens
transfers well to the task of discriminating between synthetically generated sentences and human written sentences in \textsc{DiscoSense}.%
\footnote{While \textsc{Funnel Transformer} employs the same pre-training strategy as \textsc{Electra}, we speculate that the pooling mechanism it uses to compress hidden states offsets the benefits it receives from its pre-training strategy on this task.}
Nevertheless, the fact that 
it only achieves an accuracy of 65.87\% is indicative of the challenges {\sc DiscoSense} has for existing LMs.
Note that this accuracy is much lower than those achieved by LMs on many commonsense reasoning benchmarks (see Table~\ref{tab:status_difficulty_benchmark}). These results suggest that {\sc DiscoSense} is a challenging benchmark for state-of-the-art LMs. 

Finally, we report human performance in the last row of Table~\ref{tab:quantitative_main_results}. Details of how these numbers are obtained are discussed in Section~3.4. As can be seen, the accuracy achieved by the best baseline, \textsc{Electra}, lags behind that of humans by nearly 30\% points.
\subsection{Quantitative Error Analysis}
We perform a quantitative error analysis of our best-performing model, \textsc{Electra}.
Specifically, we compute for each discourse connective the percentage of examples in the \textsc{DiscoSense} test set that are misclassified by \textsc{Electra}, with the goal of gaining a better understanding of the discourse connectives that are perceived as easy as well as those that are perceived as difficult as far as commonsense reasoning is concerned.

\begin{figure}[!t]
\centering
    \includegraphics[height=0.8\linewidth, scale=0.14]{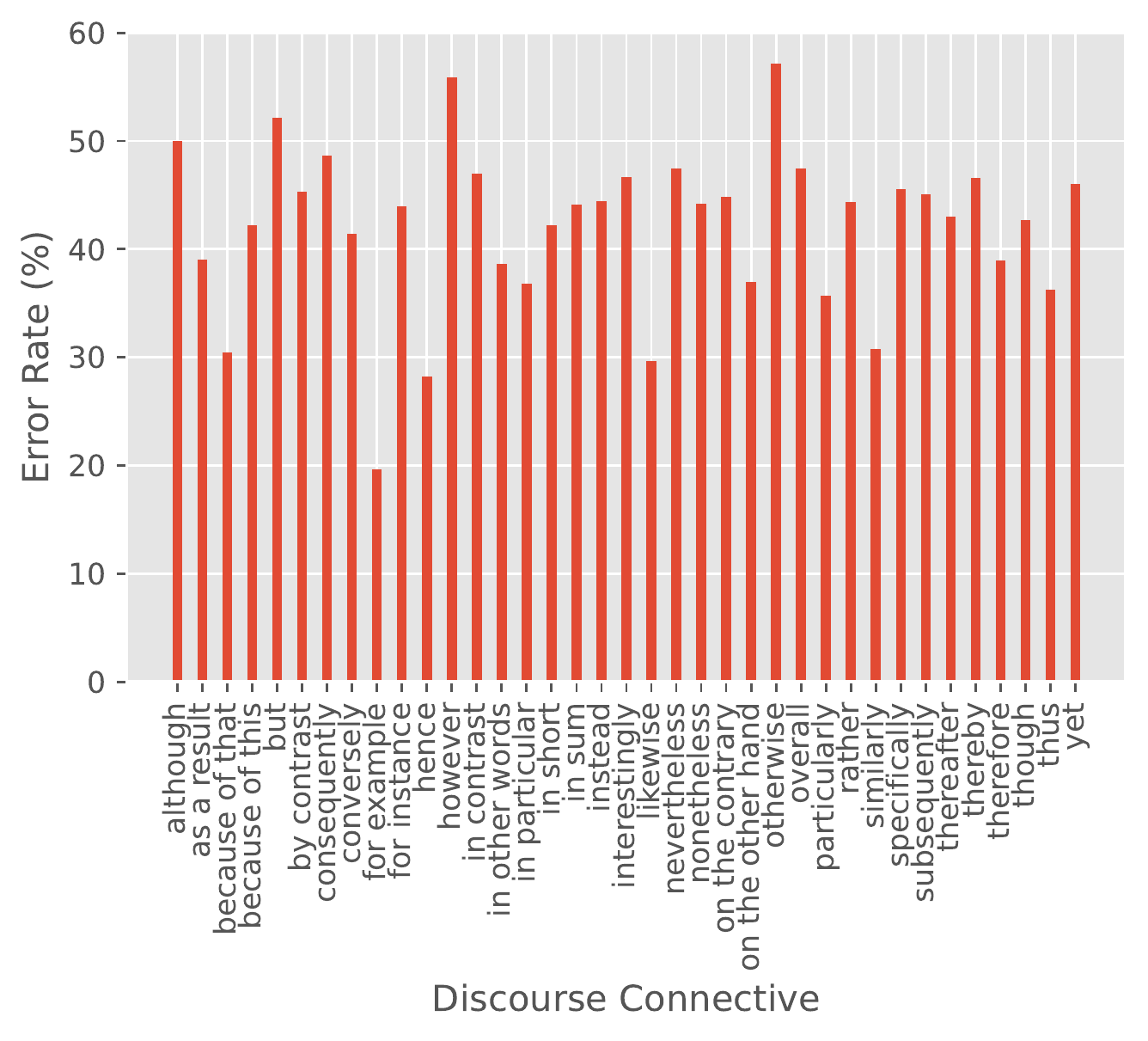}
  \caption{Error rate of \textsc{Electra} on each discourse connective in the \textsc{DiscoSense} test set.} 
\label{fig:training_marker_stats1}
\end{figure}

Results are shown in \autoref{fig:training_marker_stats1}. As we can see, the misclassification rates are highest for those discourse connectives that
express contrast (e.g., ``otherwise", ``however", ``but", ``although"). A plausible explanation for this result is that it is often hard to anticipate what a human would have in mind if they are trying to indicate the opposite of what they mean to say. On the other hand, the model finds it easy to predict sentences where the discourse connective signals compliance and exemplification (e.g., ``similarly", ``likewise", ``hence", ``because of that", ``for example").

\subsection{Qualitative Error Analysis}

\comment{
Next, we perform a qualitative error analysis of 
\textsc{Electra}
by examining the discourse connectives that are the most and the least frequently associated with the correctly classified examples on the \textsc{DiscoSense} test set. 
We found that most of the misclassified examples 
have discourse connectives that express {\em contrast} (e.g., "otherwise", "however").
A plausible explanation for this observation is that it is often difficult to anticipate what a human has in mind when s/he is trying to indicate the opposite of what s/he means to say. On the other hand, the model finds it easy to predict sentences involving discourse connectives that indicate compliance, causality, and exemplification (e.g., 
"similarly", 
"because of that", "for example").
}

To better understand the mistakes made by \textsc{Electra}, we manually inspected 100 randomly selected examples that are misclassified and identified four major reasons why they are misclassified. 

\begin{figure}[t]
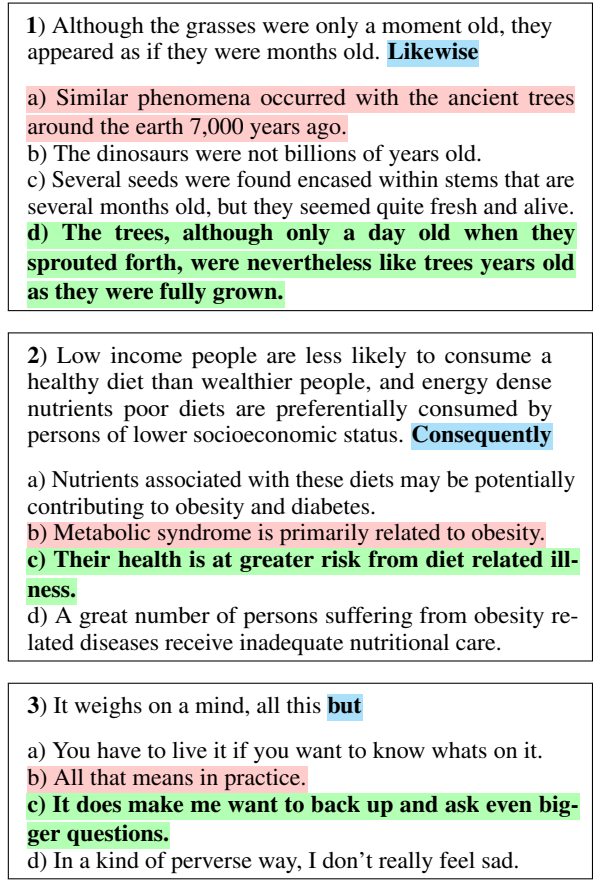

\centering\small 
{\FrameSep1pt
    \begin{framed}\small\begin{tabular}{@{}l @{}}
\aquestion{}{\textbf{1}) Although the grasses were only a moment old, they appeared as if they were months old. \discourse{Likewise}}{\incans{a) Similar phenomena occurred with the ancient trees around the earth 7,000 years ago.}}{b) The dinosaurs were not billions of years old.}{c) Several seeds were found encased within stems that are several months old, but they seemed quite fresh and alive. }{\correctans{d) The trees, although only a day old when they sprouted forth, were nevertheless like trees years old as they were fully grown.}} 
\end{tabular}
\end{framed} }

{\FrameSep1pt
    \begin{framed}\small\begin{tabular}{@{}l @{}}
\aquestion{}{\textbf{2}) Low income people are less likely to consume a healthy diet than wealthier people, and energy dense nutrients poor diets are preferentially consumed by persons of lower socioeconomic status. \discourse{Consequently}}{a) Nutrients associated with these diets may be potentially contributing to obesity and diabetes.}{\incans{b) Metabolic syndrome is primarily related to obesity.}}{\correctans{c) Their health is at greater risk from diet related illness.}}{d) A great number of persons suffering from obesity related diseases receive inadequate nutritional care.}
\end{tabular}
\end{framed} }
{\FrameSep1pt
    \begin{framed}\small\begin{tabular}{@{}l @{}}
\aquestion{}{\textbf{3}) It weighs on a mind, all this \discourse{but}}{a) You have to live it if you want to know whats on it.}{\incans{b) All that means in practice.}}{\correctans{c) It does make me want to back up and ask even bigger questions.}}{d) In a kind of perverse way, I don't really feel sad.}
\end{tabular}
\end{framed} }
\caption{Examples 
misclassified by \textsc{Electra} 
(misclassified options in \incans{pink}; ground truths in \correctans{green}).}
\label{tab:qualitative_examples}
\vspace{-2mm}
\end{figure}

    \paragraph{1.\ Less plausible endings.} This category contributes to 21\% of the errors where the model chooses a less plausible ending. Choosing a less plausible option could be associated with a partial understanding of the context or unwarranted assumptions.
In Example 1 of \autoref{tab:qualitative_examples}, 
the model makes the assumption that whatever is applicable to grass is also applicable to trees.
However, the option it ends up picking is non-factual in nature because of the phrase ``7000 years ago".

    
    \paragraph{2.\ Abstract associations.} 14\% of the errors are made due to the formation of abstract associations between concepts. The model seems to rely on certain spans of context for classification rather than understand the semantics in its entirety.
    In Example 2 of \autoref{tab:qualitative_examples}, 
    the model seems to wrongly associate ``energy dense nutrients" with ``obesity" and fails to understand that the context is discussing the correlation between nutrient deficit diet and people belonging to lower income groups.
    
    \paragraph{3.\ Complex Context Understanding.} 23\% of the examples are misclassified due to the fact that a deeper than usual reasoning is needed to understand the context. In Example 3 of~\autoref{tab:qualitative_examples}, we see that the context is about something weighing on a mind, indicating that the author may be faced with a pressing situation. The connective ``but" indicates that while the situation being dealt with is problematic or stressful, the author would still pursue it, making option c) the most plausible. Here, the model fails to understand what it means to have something weighing on mind and what that can make a person do, in this case, ``ask bigger questions".
    
    \paragraph{4.\ Lack of understanding of the discourse connective.} 
    In many cases it is difficult to pinpoint the reason why an example is misclassified. Hence, if a misclassified example is not covered by any of the first three categories, we attribute the mistake to a lack of understanding of the discourse connective.  
    This category contributes to 42\% of the errors. 
    
\subsection{Role of Context and Discourse connective}
To better understand the role played by the context and the discourse connective in a LM's reasoning process, we conduct two ablation experiments.
In the first experiment, we remove the discourse connective, so only the context and the endings are available to the LMs.
In the second experiment, we strip the context {\em and} the discourse connective, exposing only the endings to the LMs. 

Results of these experiments are shown in the \textsc{C+E} column and the \textsc{E} column of \autoref{tab:role_of_context} respectively. For comparison purposes, the results obtained by not removing anything
are shown in the \textsc{C+D+E} column.
As can be seen, when the discourse connective is removed, performance drops for all baselines except for {\sc Bert-base} and {\sc Longformer}, and when both the discourse connective and the context are removed, performance drops for all baselines.
In the case of the best baseline, \textsc{Electra}, performance drops abruptly as information is withdrawn (\textsc{C+E}: $17.91\%$ and \textsc{E}: $44.27\%$), thus highlighting its reliance on both pieces of information for its competitive performance. Overall, these results suggest that reasoning over both the context and the connective is necessary for this task.
%
It is worth mentioning, though, that even when both the context and the connective are removed, all the LMs still manage to achieve an accuracy of more than 30\%. Additional experiments are needed to determine the reason why
they perform considerably better than random guess when only the endings are given.
This will most likely involve an examination of whether there are systematic differences between the human-generated sentences and their automatically generated counterparts.

\begin{table}[t]
\footnotesize
\begin{center}
\begin{tabular}{lp{1.3cm}p{1.3cm}p{1.3cm}}
\toprule
\bf Model & \bf C+D+E & \bf C+E & \bf E\\ 
\midrule
\textsc{Bert-base} & 32.86 / 0.5 & 33.95 / 0.8 & 32.27 / 1.3 \\
\textsc{Bert-large} & 34.25 / 1.0 & 33.30 / 0.6 & 30.29 / 0.9\\
\textsc{Roberta-base} & 34.11/ 0.5 & 32.73 / 0.4 & 32.49 / 0.7\\ 
\textsc{Roberta-large}  & 34.99 / 0.2 & 34.94 / 0.9 & 32.70 / 0.7\\
\textsc{Albert-xxl}  & 50.91/ 1.4  & 50.87 / 1.2 & 38.04 / 0.4\\
\textsc{Longformer}  & 35.29 / 0.8 & 36.82 / 0.8 & 33.18 / 1.0\\
\textsc{XLNet Large} & 36.71 / 0.8 & 36.68 / 1.1 & 31.87 / 0.4\\
\textsc{Funnel-XL} & 35.22 / 1.9 & 34.62 / 4.9 & 30.76 / 4.2\\
\textsc{Electra-Large} & 65.87 / 2.3 & 56.75 / 0.8 & 43.33 / 2.2\\
\bottomrule
\end{tabular}
\end{center}
\caption{Accuracies (best results among 8 epochs when averaged over 5 runs with random seeds) on the \textsc{DiscoSense} test set with specific pieces of information from the input removed. D, C, and E refer to the discourse connective, the context and the endings respectively. The numbers following '/' are the standard deviations.}
\label{tab:role_of_context}
\vspace{-2mm}
\end{table}

\comment{
\begin{table}[t]
\footnotesize
\begin{center}
\begin{tabular}{p{4cm}p{1.5cm}p{1.5cm}}
\toprule
\bf Model & \bf C+E & \bf E\\ 
\midrule
\textsc{Bert-base} & 33.95 / 0.8 & 32.27 / 1.26 \\
\textsc{Bert-large} & 33.30 / 0.6 & 30.29 / 0.93\\
\textsc{RoBERTa-base} & 32.73 / 0.36 & 32.49 / 0.74\\ 
\textsc{Robert-large}  & 34.94 / 0.92 & 32.70 / 0.65\\
\textsc{ALBERT-xxlarge-V2}  & 50.87 / 1.2 & 38.04 / 0.42\\
\textsc{Longformer Base}  & 36.82 / 0.82 & 33.18 / 0.97\\
\textsc{XLNet-Large} & 36.68 / 1.05 & 31.87 / 0.29\\
\textsc{Funnel-Transformer-XL} &  34.62 / 4.85 & 30.76 / 4.21\\
\textsc{Electra-Large} & 56.75 / 0.84 & 43.33 / 2.07\\
\bottomrule
\end{tabular}
\end{center}
}



\subsection{\textsc{DiscoSense} for Transfer Learning}
Next, we look at \textsc{DiscoSense} from the perspective of a transfer learning source. Specifically, to understand whether fine-tuning a LM on \textsc{DiscoSense} 
can improve its performance on a related dataset, \textsc{HellaSwag}, we perform sequential fine-tuning, where we fine-tune each baseline LM on the \textsc{DiscoSense} training set followed by the \textsc{HellaSwag} training set (both for 4 epochs). Note that discourse connectives are removed from the input because \textsc{HellaSwag} does not have them. 

Results 
on the {\em validation} split of \textsc{HellaSwag} are shown in Table~\ref{tab:seq_fintuning}. Specifically, the HS column shows the results of the baselines on \textsc{HellaSwag} and the DS$\rightarrow$HS shows the results of the baselines after sequential fine-tuning. As we can see, sequential fine-tuning yields performance improvements with almost all LMs. 
Notably, the improvement is more pronounced for \textsc{Electra} (4.3\%) than for \textsc{Albert} and the \textsc{Bert}-based models. One plausible reason is that \textsc{Electra} does not struggle as much in understanding \textsc{DiscoSense} as the \textsc{Bert}-based models do, and as a result, it shows a bigger improvement, possibly benefitting from the diverse contextual nature of \textsc{DiscoSense}.

Finally, we evaluate the sequentially fine-tuned \textsc{Electra-large} model on the \textsc{HellaSwag} {\em test} split. 
The model achieves an accuracy of 90.76\%, considerably outperforming its vanilla fine-tuning counterpart (85.75\%) and only underperforming 
models that have 4x (e.g., \newcite{He2021DeBERTaDB}, \newcite{Lourie2021UNICORNOR}) and 32x more parameters (e.g., \newcite{Lourie2021UNICORNOR}) and are trained on 23x more data. 

\hspace{-5mm}
\begin{table}[t]
\footnotesize
\centering

\begin{tabular}{lccc}
\toprule
\bf Model & \bf HS & \bf DS$\longrightarrow$HS \\ 
\midrule
\textsc{Bert-base-uncased} & 38.47 & \textbf{40.38}\\
\textsc{Bert-large-uncased} & 44.36 & 42.54\\
\textsc{Roberta-base} & 58.21 & 57.00\\
\textsc{Roberta-large} & 81.50 & \textbf{82.34}\\
\textsc{Albert-xxlarge-V2} & 80.97 & \textbf{81.47}\\
\textsc{Xlnet-large} & 76.47 & \textbf{76.56}\\
\textsc{Electra-Large} & 86.90  & \textbf{91.50} \\
\textsc{Funnel-Transformer-XL} & 86.88 & \textbf{87.50}\\
\bottomrule
\end{tabular}
\caption{Results of sequential fine-tuning 
on the validation split of \textsc{Hellaswag}.}
\label{tab:seq_fintuning}
\vspace{-2mm}
\end{table}

\section{Conclusion}
\label{sec:conclusion}
Motivated in part by the fact that existing pre-trained language models have surpassed human performance on numerous commonsense reasoning datasets, 
we introduced \textsc{DiscoSense}, 
a challenging benchmark that concerns commonsense reasoning with 
discourse connectives
to determine the most plausible ending of a sentence.
This task was made difficult 
by the synthesis of high quality complex examples, which was made possible through coupling highly competitive conditionally trained models for language generation with Adversarial Filtering. 
The best performing 
model on \textsc{DiscoSense} only 
achieved an accuracy of $65\%$, significantly lagging behind humans. 
This makes \textsc{DiscoSense} an ideal benchmark for next-generation commonsense reasoning systems.

\section*{Ethical Considerations}
Following the guidelines in \newcite{48120}, \newcite{bender-friedman-2018-data}, and \newcite{10.1145/3458723}, we believe that we have provided all the necessary information in our description of \textsc{DiscoSense}. In this section, we focus on ethical considerations.


\textbf{Bias mitigation.} While it may not be possible to eliminate all of the biases that exist in a dataset, we have certainly taken steps to mitigate biases in {\sc DiscoSense}.
Adversarial Filtering has been shown to be an effective de-biasing approach to remove annotation artifacts, and we have taken a step further to improve this approach through conditional text generation. In addition, to our knowledge, our work has used more capable generators and discriminators (adversarial filter) to synthesize text in comparison to other works \cite{zellers-etal-2018-swag, zellers-etal-2019-hellaswag, DBLP:conf/icml/BrasSBZPSC20}.

\textbf{Human annotator information.} All annotators/verifiers were hired during Summer 2021 as student workers (20-25 hours/week) with full consent. All of them were undergraduate and graduate students aged around 20-24. The group comprised both male and female students with members belonging to different ethnicity, namely Asian, Caucasian, and Hispanic. All annotators were native English speakers. Additional details on the selection of annotators can be found in Appendix~A.1. The annotators were 
compensated with a hourly rate of 10 US dollars.


\textbf{Steps taken to protect annotators from harmful content.} All annotators were provided with a thorough instructional training session in which they were instructed on how to annotate the data, how to go about the whole task, 
and what kind of examples to skip. Before we shared the data, we performed filtering of examples based on sensitive/offensive keywords. 
After the filtering process, 
we provided 
the annotators with a document that contains instructions on how to annotate and how to go about the whole task (see Appendix~A.2). They were asked to follow their own pace (the amount of time they can spend per example). They were asked to attempt examples that were specifically related to commonsense reasoning tasks. Since the aforementioned keyword-based approach for filtering harmful content may not be able to identify all harmful/offensive documents, the annotators were provided with an opportunity to skip examples that they would consider offensive, sensitive or challenging enough to confuse them.

\textbf{Is this dataset consistent with the terms of use and the intellectual property and privacy right of people?} 
The most important term of use for this dataset is that it shall primarily be used for NLP research. The source text of this dataset was obtained from \textsc{Discovery} and \textsc{Discofuse}, both of which have been there for a long time. These datasets have been obtained from Common Crawl and Wikipedia data, which is public information. Therefore these data sources do not contain any information that is non-public. We agree with the authors of \textsc{Discovery} and \textsc{Discofuse} that these data sources do not seem to mis-represent any community nor can be used to identify a certain set of individual known outside public information. Through conditional text generation, which has been used to synthesize commonsense knowledge text with discourse connectives, we present text that is mostly suitable for commonsense reasoning tasks, making this work consistent with the terms of use. We believe that our work does not have use cases that would usually be considered out-of-bounds for NLP research.

\textbf{Is there anything about the composition of the dataset or the way it was collected and pre-processed/cleaned/labeled that might impact future uses?} We have highlighted all the necessary information required by the user of this dataset to use it for their own use case. Each example has  gone through multiple rounds of screening. We do not expect to see any risk being posed by the user of this dataset nor any financial harm associated with its use.

\textbf{Will the dataset be distributed to third parties outside of the entity (e.g., company, institution, organization) on behalf of which the dataset was created?} We will open-source all the models and data produced from this work immediately after publication. We plan to release it on a GitHub repository with the MIT license and also make it available on Hugging Face Datasets.

\textbf{If others want to extend/augment/build on/contribute to the dataset, is there a mechanism for them to do so?}
We have provided all the essential information needed by a user to extend this work. They will have access to the data and the models which they can use for experimentation. We will continue to monitor the GitHub repository to resolve issues.


\section*{Limitations}

Next, we discuss some limitations of our work.

{\bf Limitations of (Conditional) Adversarial Filtering.} 
Recall that we seek to create a challenging commonsense reasoning benchmark by automatically removing annotation artifacts%
\footnote{An example of an annotation artifact would be that examples in a commonsense reasoning task can be solved by a pre-trained language model using unintended artifacts that exist in the data such as lexical overlap/similarity.} and replacing easy options in examples via the design and use of Conditional Adversarial Filtering (CAF). Although CAF is an improved version of Adversarial Filtering (AF), which has been frequently used in the last few years in the construction of commonsense reasoning benchmarks, it is not without its limitations. Specifically, how well CAF can identify annotation artifacts and easy options and subsequently remove artifacts and replace easy options with difficult ones depends on how good the discriminator and the generators are. As discussed before, while the discriminator and generators we use in the creation of \textsc{DiscoSense} are stronger than those used in the creation of virtually all other commonsense reasoning benchmarks (e.g., \textsc{Swag} and \textsc{Hellaswag}), these discriminator and generators are still not perfect. In particular, the fact that the best-performing baseline, \textsc{Electra}, achieves an accuracy that is substantially higher than random guess (i.e., 67\% vs.\ 25\%) is an indication that the discriminator and generators fail to remove all annotation artifacts and/or replace easy options with sufficiently difficult ones for state-of-the-art pre-trained language models. As pre-trained language models continue to improve, we do expect that the family of AF approaches will become more effective. Nevertheless, moving forward, researchers should think about whether there are alternative, non-AF-based approaches for creating challenging commonsense reasoning benchmarks that do not suffer from the limitations of AF-based approaches.



{\bf Coverage of discourse connectives.} While we have taken measures to ensure that \textsc{DiscoSense} has a good coverage of discourse connectives, there are still many connectives that are not present in \textsc{DiscoSense} due to the ambiguity they give rise to. For example, having ``and" does not make it clear what the next sentence should talk about given a context, meaning that it is likely for these connectives to have many endings that are equally plausible. Given budgetary constraints, we do not want to waste our human verification effort on identifying and filtering the potentially large number of examples that contain equally plausible endings as a result of these ambiguous discourse connectives. So, we have avoided their inclusion in \textsc{DiscoSense} thus far. However, to fairly evaluate how good a model is in reasoning with discourse connectives, we should augment \textsc{DiscoSense} with ambiguous discourse connectives in the future.

{\bf Types of reasoning.} Although there is a high coverage in the types of commonsense reasoning \textsc{DiscoSense} aims to study (e.g., physical world reasoning, social commonsense reasoning, numerical reasoning, linguistic reasoning, temporal reasoning, abductive reasoning), there are other kinds of reasoning studied within the NLP literature that this benchmark does not aim to evaluate upon, such as multi-hop reasoning and symbolic reasoning.  It is still not clear how adversarial approaches can be applied to make these kinds of reasoning difficult. We leave this component to a future work.

\section*{Acknowledgments}
We thank the three anonymous reviewers for their
insightful 
comments on an earlier
draft of the paper. 
This work was supported in part by NSF Grant IIS-1528037.
Any opinions, findings, conclusions or recommendations expressed in this paper are those of the authors and do not necessarily reflect the views or official policies, either expressed or implied, of the NSF.

\comment{
\section*{Ethical Considerations}
\textbf{Steps taken to protect annotators from harmful content.}
The data that has been provided to our generator LMs were taken from \textsc{Discovery}~\cite{sileo-etal-2019-mining} and \textsc{Discofuse} ~\cite{geva-etal-2019-discofuse}, both of which have been peer-reviewed by their authors. Hence, 
the source data has presumably gone through strict quality checks. In cases where the generator generated a sentence that can be considered objectionable or offensive, we used a filtration process to remove them. Our algorithmic filtration process was based on sensitive keywords. If the context sentences contained keywords from the sensitive keyword list, the whole example would be automatically removed \cite{strassel2000quality}. Moreover, annotators were asked to not solve any example that they would deem as biased/harmful. They were given the freedom to select out examples which they perceived as safe.
}


\bibliographystyle{acl_natbib}
\bibliography{acl2021,anthology}

\appendix

\section{Crowdsourcing Details}

We performed crowdsourcing to determine if humans are able to identify the ground truth in each example present in the training and test sets of \textsc{DiscoSense}. 

\subsection{Selection of Annotators}
We selected undergraduate and graduate student workers primarily on the basis of relevant background knowledge and their current skill set.  
Since the dataset features intricate English sentences, we shortlisted students who are native English speakers. We presented all shortlisted student workers with a one-hour training session followed by an evaluation test and ranked them based on their performance on the test. 
We then hired the top-performing students as our annotators and asked them to answer each example in \textsc{DiscoSense}. 
We closely monitored the progress and the performance of each annotator 
and provided timely feedback through virtual meetings on how his/her way of performing annotation can be improved by understanding the difficulties encountered by him/her and  
rectifying any problems if they have been doing the annotation in an incorrect manner. All of the annotators were made aware of how the annotated data would be used and its implications. The student workers who did not perform the annotation work in a satisfactory manner were replaced by the next best performing student on the waitlist. All student workers were paid $\$10$/hr \cite{fort2011amazon,cohen2016ethical}.

\subsection{Instructions Provided to Annotators}

Below are the instructions we presented to the annotators during our one-hour tutorial. Each of them received a copy of these instructions plus numerous examples (one of them is shown below) at the end of the tutorial and was given an opportunity to ask clarification questions about the instructions and the annotation process in general after a closer examination of these instructions and examples.

\paragraph{Motivation.}
Academic research is an exploration activity to solve problems that have not been completely solved before. By this nature, each academic research work must sit at the frontier of the field and present novelties that have not been addressed by prior works. Machine Learning (specifically Deep Learning) has seen advancements in numerous domains such as personal assistants, machine translation, etc. What you may have seen is a particular Machine Learning algorithm being deployed in finished end products, but what we do not observe are the research challenges that were overcome behind the scenes to get there. What you will be involved in is one of the research tasks that is far from being solved. You will be contributing towards a meaningful task that has the potential to make scientific advancements.
We are dealing with Commonsense Reasoning with regard to Natural Language Processing. Specifically, we are addressing the problem of how machines can learn to reason in the manner humans do with textual data. This task often requires humans to make use of the information they have acquired about our world (e.g., how the physical world works), and reason about what they read and how it complies with what they already know. The reasoning works on multiple levels: firstly we understand the information we read, make sure we understood it properly, and then reason through various means about it to ensure that it makes sense with what we know. The following quote sums it up nicely. \\ 

\fbox{\begin{minipage}{17em}
The brain is an abduction machine, continuously trying to prove abductively that the observables in its environment constitute a coherent situation. 

\vspace{2mm}
--- Jerry Hobbs, ACL 2013 Lifetime Achievement Award winner
\end{minipage}}
\newline
\\ \\ Current Deep Learning models are very good at picking unintended signals to arrive at the correct answer, but this is not what we intend them to do.
For instance, if ``not" is present within a sentence, then the model might be biased towards predicting negation even though that might not be the case.
We require them to pick the right answer 
through a reasoning process that is as close to the human reasoning process as possible.
To address this task, we need to build a dataset that aims at providing ``correct" signals to our models, and hopefully the models can learn to reason reasonably well once they are trained on this dataset.

\paragraph{High level description of the task.} 
The task description is straightforward.\\ \\
\fbox{\begin{minipage}{18em}
Given a context and a discourse connective,\\ 
predict which ending is the best to the best\\
of your knowledge/capability.
\end{minipage}}
\newline
\\

While reading the contexts, you 
should understand what the discourse connective is supposed to convey.
A discourse connective is a word/phrase 
used to connect two sentences and reveal their relationship. 
Consider the following examples.
\begin{enumerate}
    \item \textcolor{teal}{I am feeling hungry}, \textbf{as a result}, \textcolor{orange}{I cooked lunch}.
    \item \textbf{Although} \textcolor{teal}{I liked reading the book}, \textcolor{orange}{there were some major flaws}.
\end{enumerate}

In these examples, the green-colored text is the context and the discourse connective is boldfaced. Notice how the text in orange is framed in accordance with the discourse connective. For instance, if ``as a result" is present, then the ending will most likely be about the consequence of what is described in the context; in contrast, if ``although" is the discourse connective, then the ending needs to take a contrasting standpoint. Hence, the discourse connective decides what the ending needs to talk about.
We have provided a list of the discourse connectives you can expect in the ``Role of discourse connectives" section. You are required to be completely familiar with the role each
connective is supposed to play.

Figure~6 shows an example you might expect in the dataset.
Any ending that violates what we know about how the physical world works or challenges our notions about anything in particular needs to be discarded.
You have to choose the most plausible ending, which is the ending that is the most feasible amongst four endings. In cases where two or more endings seem equally feasible, the following criteria should be used.

\begin{figure}[t!]
\centering
\footnotesize
\begin{subfigure}[b]{0.475\textwidth}
\centering
    {\FrameSep1pt
    \begin{framed}\small\begin{tabular}{@{}l @{}}
\aquestion{}{I love that he’s able to use wired as a venue for launching future
bestsellers \discourse{though}}{a) I think the wired article is a bit too long.}{b) I don’t think its a bad thing for him to do so.}{c) I do agree with some of the other reviews that wired is not a very well written book.}{\correctans{d) Honestly, I might have preferred the podcast of his presentation on the topic.}}
\end{tabular}
\end{framed} }
\end{subfigure}

\comment{
\begin{subfigure}[b]{0.475\textwidth}
{\FrameSep1pt
    \begin{framed}\scriptsize\begin{tabular}{@{}l @{}}
\aquestion{}{\textbf{2.} Dependents are not generally permitted to accept employment. \discourse{Because of this}}{\correctans{a) Students may find that they cannot support a family while
              studying at the university and must be prepared to leave family
              in the home country.}}{b) They cannot be rehired for another position within this company.}{c) The company is unable to recruit new employee at this time.}{d) They are often forced to seek work in the underground economy.}
\end{tabular}
\end{framed} }
\end{subfigure}
}
\caption{Example taken from the \textsc{DiscoSense} training set. The correct answer is {\bf boldfaced}.} 
\label{fig:teaser_examples}
\end{figure}

\begin{itemize}
    \item Any ending that seems to be indisputably correct can be regarded as the best. 
    \item The best ending will not be inconsistent with the context and the discourse connective. 
    \item If all three options seem incorrect and implausible and the remaining option makes the most sense relatively, then it should be chosen since the correct ending is the best ending by default.
    \item Two endings can make sense, but the ending that is likely to be more sensible is what you should consider as the best. For example:
    \begin{itemize}
        \item Context: A man is singing into a microphone.
        \item Ending 1: A man performs a song.
        \item Ending 2: A man is performing on stage.
    \end{itemize}
Ending 2 is not incorrect but ending 1 makes more logical sense to us as humans. In such cases, mark what seems more logical to you.
\end{itemize}

\paragraph{Role of discourse connectives.}
Please make sure that you understand the role of each connective.
\begin{itemize}
\item although: in spite of the fact that; even though
\item as a result: because of something
\item because of this: for the reason that
\item because of that: for the reason that
\item but: used for joining two ideas or statements when the second one is different from the first one
\item by contrast: used to express difference with something
\item consequently: as a result
\item conversely: introducing a statement or idea which reverses one that has just been made or referred to
\item for example: used to introduce something
\item for instance: as an example
\item hence: as a consequence; for this reason
\item however: used to introduce a statement that contrasts with or seems to contradict something that has been said previously
\item in contrast: used to express difference with something
\item in other words: to put it another way
\item in particular: especially; specifically
\item in short: to sum up; briefly
\item in sum: to sum up; in summary
\item instead: as an alternative or substitute
\item interestingly: in a way that arouses curiosity or interest
\item likewise: in the same way
\item nevertheless: in spite of that; notwithstanding; all the same 
\item nonetheless: in spite of that; nevertheless
\item on the contrary: conversely; used to intensify a denial of what has just been implied or stated by suggesting that the opposite is the case
\item on the other hand: used to introduce a contrasting point of view, fact, or situation
\item otherwise: in circumstances different from those present or considered; or else
\item overall: taking everything into account
\item particularly: especially
\item rather: used to indicate one's preference in a particular matter; preferably
\item similarly: in a similar way
\item specifically: in a way that is exact and clear; precisely
\item subsequently: after a particular thing has happened; afterward 
\item thereafter: after that time
\item thereby: by that means; as a result of that
\item therefore: for that reason; consequently
\item though: despite the fact that; although
\item thus: as a result or consequence of this; therefore
\item yet: so far; up until the present or a specified or implied time; by now or then
\end{itemize}

\end{document}